\documentclass[9pt]{IEEEtran}
\usepackage{graphicx}

\usepackage{fancyvrb}
\fvset{numbers=left,fontsize=\footnotesize}
\DefineShortVerb{\|}

\usepackage{paralist}

\begin{document} 

\title{Shadows and Headless Shadows: an Autobiographical Approach to Narrative Reasoning}
\author{
\IEEEauthorblockN{Ladislau B{\"o}l{\"o}ni}\\
\IEEEauthorblockA{
Dept. of Electrical Engineering and Computer Science\\
University of Central Florida\\
Orlando, FL 32816--2450\\
lboloni@eecs.ucf.edu
}
} 
\maketitle

\sloppy

\begin{abstract}
\begin{quote}

The Xapagy architecture is a story-oriented cognitive system which relies exclusively on the autobiographical memory implemented as a raw collection of events. Reasoning is performed by {\em shadowing} current events with events from the autobiography. The shadows are then extrapolated into {\em headless shadows} (HLSs). In a story following mood, HLSs can be used to track the level of surprise of the agent, to infer hidden actions or relations between the participants, and to summarize ongoing events. In recall mood, the HLSs can be used to create new stories ranging from exact recall to free-form confabulation. 

\end{quote}
\end{abstract}

\section{Introduction}

The Xapagy cognitive architecture has been designed with the explicit goal of {\em narrative reasoning:} to model and mimic the activities performed by humans when witnessing, reading, recalling, narrating and talking about stories. Xapagy has been developed from scratch, which required us to revisit many of the problems identified in the classic literature of the story understanding. 

In particular, the Xapagy architecture takes an unusual approach to knowledge representation: the {\em autobiographical narrative is the only source of knowledge}, the {\em autobiographical memory is the only memory model} and there is {\em no retrieval from long term into working memory}. The claim made by this paper is that these design decisions, supported by the shadowing / headless shadows based reasoning mechanism, can yield a system which can successfully perform narrative reasoning. 

We support the claim by a detailed description of the representation and reasoning model. Throughout the paper, we will show a number of snippets in the Xapi language, used by Xapagy for the external representation of stories. The majority of these examples refer to the world of Little Red Riding Hood (LRRH), a popular children's story, often used as example in story understanding systems. These snippets, including a full translation of LRRH into Xapi can be downloaded from the Xapagy website together with a runnable version of the Xapagy agent.

The reminder of the paper is organized as follows. Section~\ref{sec:XapiIntro} introduces the basic components of Xapagy and the Xapi language. Section~\ref{sec:FocusAndShadows} discusses the focus and the shadowing mechanism while  Section~\ref{sec:HeadlessShadows} discusses the mechanism of creating headless shadows (HLSs) and associated structures. Finally, Section~\ref{sec:NarrativeReasoning} shows how HLSs are used to used to infer surprise, continuation of stories as well as the discovery of missing action, missing relations and summarizations. We compare Xapagy with related systems in Section~\ref{sec:RelatedWork} and we conclude in Section~\ref{sec:Conclusions}.  
\section{An introduction to Xapagy and the Xapi language}
\label{sec:XapiIntro}

%
%

\subsection{The elements of the Xapi language}

We start the introduction with an informal presentation of the Xapi language. We position Xapi as a {\em pidgin language} which, while not matching the expressiveness of a natural language, can however, be used as a means of communication between Xapagy agents and humans.

The Xapagy agent maintains a {\em focus} of several {\em instances}. Every instance is a member of a single {\em scene}. Instances are characterized by a collection of attributes, represented as an overlay of {\em concepts}. A {\em concept overlay} is the simultaneous activation of several concepts with specific levels of energy. Instances can acquire new concepts, but they never loose concepts. Concepts have {\em area}, they can {\em overlap} each other, and they can {\em impact} each other when added to an overlay

The Xapagy agent evolves through the instantiation of {\em verb instances} (VIs). The most important part of the VI is the verb part which is represented by a {\em verb overlay}. Verbs are similar to concepts, with the difference that certain verbs have {\em side-effects} which modify the focus at their instantiation. Other VI parts can be instances or concept overlays, depending on the type of the VI. 

The Xapi language is a way to describe a sequence of VIs. A Xapi sentence is a collection of sentence parts separated by ``/'' and terminated with a period ``.'' or question mark ``?''. Different Xapi sentence forms correspond to the specific VI types. Usually, one Xapi sentence is converted to a single VI. In order to make writing stories easier, Xapi also supports several constructs where a single Xapi sentence might be translated into several VIs. 

The sentence parts are described through {\em verb words} and {\em concept words}. Words are mapped to overlays by the {\em dictionary} of the agent. For instance, the word ``girl'' maps into the concept overlay |[young female human]|. The dictionary and the concept database form the {\em domain knowledge} of the agent. Different agents might have different domain knowledge - thus the meaning of the word might differ between agents. 

Xapagy supports three simple and one composite VI types, with the corresponding Xapi sentences. 

\noindent In a {\bf subject-verb-object (SVO)} sentence the subject and the object are instances:

\begin{quote}
\begin{Verbatim}
The girl / hits / the wolf.
\end{Verbatim}
\end{quote}
   
The article ``the'' shows that we are referring to an instance already existing in the focus. An instance can be created by simply referring to it by the ``a/an'' article instead of ``the''. If there was no wolf in the focus, we can say:

\begin{quote}
\begin{Verbatim}
The girl / hits / a wolf.
\end{Verbatim}
\end{quote}

In this type of simple reference, we need to provide sufficient attributes for the subject and the object to be able to distinguish them in the current scene in the focus. If we have two wolfs in the focus, we might need to say:

\begin{quote}
\begin{Verbatim}
"LRRH" / hits / the big wolf.
\end{Verbatim}
\end{quote}

In these examples, S-V-O VIs have been used to represent an {\em action} in time. S-V-Os can also be used to represent a {\em relation} between two instances:

\begin{quote}
\begin{Verbatim}
"LRRH" / loves / "Grandma". 
\end{Verbatim}
\end{quote}

\noindent A {\bf subject-verb (S-V)} sentence is similar to an S-V-O sentence without the object part. The subject is resolved as before:

\begin{quote}
\begin{Verbatim}
The girl / cries.
\end{Verbatim}
\end{quote}

\noindent A {\bf subject-verb-adjective (S-V-Adj)} sentence contains a subject which is an instance and an adjective which is represented as a concept overlay. S-V-Adj VIs can be used to represent the acquisition of a new attribute by an instance through the |is-a| verb:

\begin{quote}
\begin{Verbatim}
"LRRH" / is-a / young girl.
\end{Verbatim}
\end{quote}

Another use, specific to Xapagy, is when an instance {\em changes} into another instance, with a new set of incompatible attributes:

\begin{quote}
\begin{Verbatim}
The wolf / changes / dead.
\end{Verbatim}
\end{quote}

Xapagy has a single composite VI, the {\bf quote}. In the quote-type VI, the subject performs a quoting action, specifying a scene and the {\em inquit} VI which takes place in that scene. 

\begin{quote}
\begin{Verbatim}
"Grim" / says in "ChildrenStory" // 
   the wolf / swallows / the girl.
\end{Verbatim}
\end{quote}
 
Beyond proper quotations, quote VIs are also used to represent orders, plans, conversations, questions about hypotheticals and so on.

%
%

 \subsection{Identity relation}
 
Xapagy uses multiple instances where colloquial speech identifies only one entity. These instances are then connected through the {\em identity relation}. For instance, the alive wolf and the dead wolf are represented by two different instances (as the attributes of the dead wolf can not be acquired from the attributes of the alive animal with simple addition of concepts). So the statement

\begin{quote}
\begin{Verbatim}
The wolf / changes / dead.
\end{Verbatim}
\end{quote}

\noindent will actually create a new instance, also labeled as wolf, which will be connected through an identity relationship to the dead wolf. 

\begin{quote}
\begin{Verbatim}
The wolf alive / is-identical / the wolf not-alive.
\end{Verbatim}
\end{quote}

In this case, both wolf instances are in the same scene (although the alive instance will be expired from the focus). In other occasions, the instances connected through identity relations are in different scenes. For instance, LRRH recalling her youngful adventures to her children, is connected to the (younger) LRRH in the forest with an identity relation:

\begin{quote}
\begin{Verbatim}
The old "LRRH" / is-identical / 
   the young "LRRH" --in-- scene "Forest".
\end{Verbatim}
\end{quote}

%
%

 \subsection{Reference types}

Xapi allows several ways to refer to an instance. The simplest is {\em reference by attribute} in the current scene:

\begin{quote}
\begin{Verbatim}
The young girl / cries.
\end{Verbatim}
\end{quote}

The words ``young girl'' represent a set of attributes in the form of a concept overlay. Xapi will look in the current scene for an instance which has attributes compatible with this concept overlay. If multiple compatible instances are present, the parser uses a combination of the match score and the strength of the instances in the focus to decide on the reference. 

The scene in which the attributes are resolved are the current scene for S-V-O, S-V and S-V-Adj and the subject of a QUOTE. For the inquit of the QUOTE type VI, the attributes are resolved in the scene specified in the quote.

\begin{quote}
\begin{Verbatim}
"LRRH" / says in "Forest" // 
   "LRRH" / is-a / young.
\end{Verbatim}
\end{quote}

In this sentence, the narrating old LRRH, specifies that the instance of LRRH she is talking about in the scene Forest, is young. 

Reference by attribute is not always possible because we might need to refer to instances which are not distinguished by any specific attributes. In this case we can use {\em reference by relation} by specifying a base instance, then a series of relations and attribute based references. For instance, we can refer to the eye of the wolf:

\begin{quote}
\begin{Verbatim}
The eye -- of -- wolf / is-a / big. 
\end{Verbatim}
\end{quote}

\noindent where the word ``of'' selects the ownership relation. Reference by relation can be also used to refer to instances not in the default scene, using the scene membership relation |-- in --| followed by the reference to the scene. 

\medskip

A special type of reference applies to {\em enumerated groups} which can be referred to by adding the ``+'' sign between the members:

\begin{quote}
\begin{Verbatim}
The wolf + hunter / are-fighting. 
\end{Verbatim}
\end{quote}

Note that this is an S-V type VI, with the subject being a single instance, which happens to be a group. The group instance is created automatically in the focus when first referred to.

%
%

 \subsection{Macro statements}
 
\begin{figure*}[t!]
\begin{center}
\includegraphics[scale=0.35]{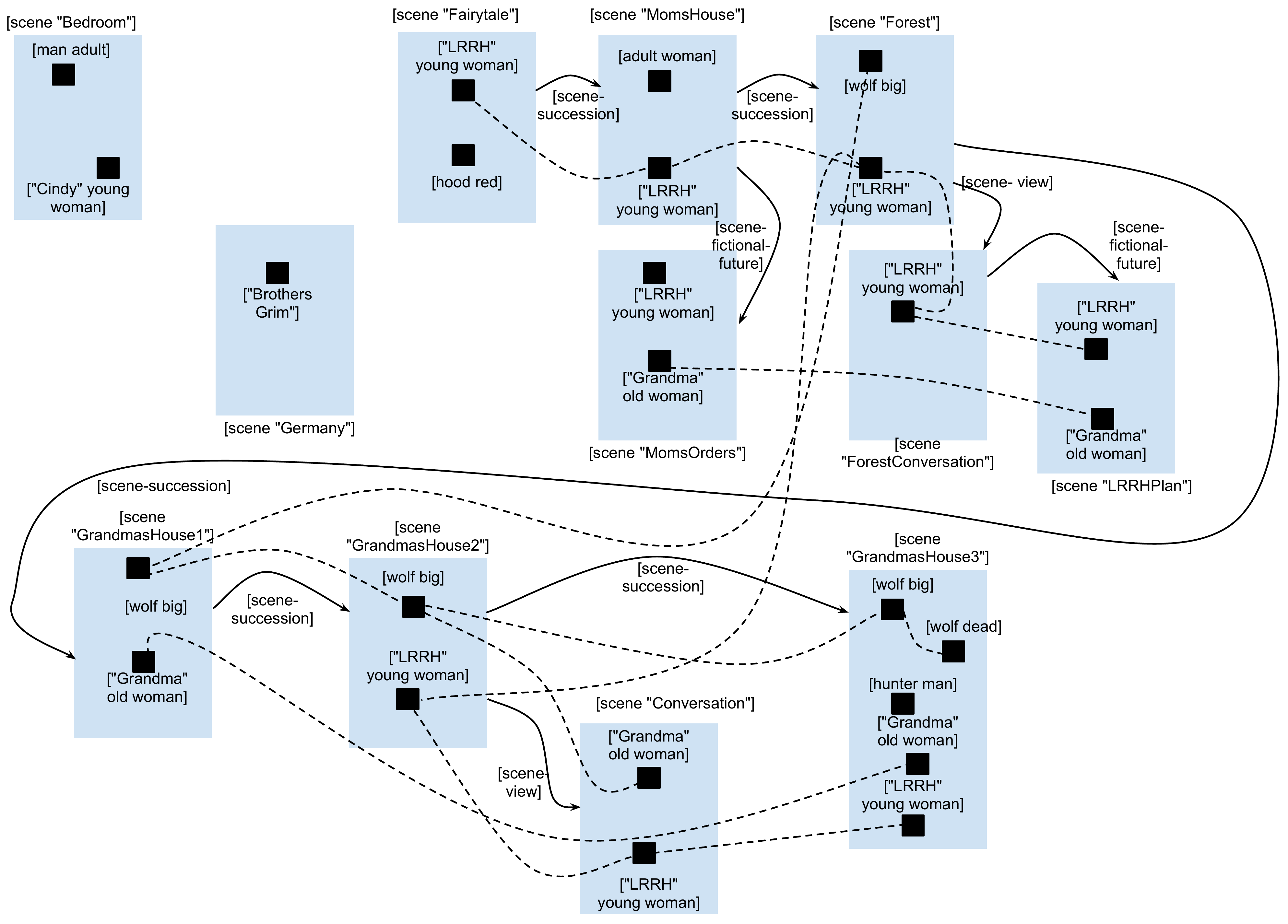} 
\caption{\label{fig:LRRHFull} The scene and identity structure of the story of Little Red Riding Hood written in the
Xapi pidgin and parsed into Xapagy. We ignored some of the entities present in the story (hood, basket and so on), instead focused on the main characters.}
\end{center}
\end{figure*}

In practice, when writing Xapagy stories, we frequently need to set up new scenes, create a set of instances, and connect them by identity relations to the existing scenes. A longer story requires the agent to create a many scenes, and the characters in the story can be represented through a number of instances. For instance, one translation of the story of LRRH into Xapi created 10 separate instances of LRRH (see Figure~\ref{fig:LRRHFull} for a simplified diagram - for a full trace, we refer the reader to {\tt http://www.xapagy.com/?page\_id=6}). 

To simplify the writing of longer stories, Xapi supports macro statements which simplify writing repetitive sequences of VIs, for instance, the initialization of scenes: 

\begin{quote}
\begin{Verbatim}
$NewSceneCurrent "Conversation", view, 
  little girl "LRRH" -> "LRRH", 
  old woman "Grandma" -> wolf
\end{Verbatim}
\end{quote}

In this example, we created a new scene ``Conversation'', connected with a view relationship to the current scene. In this scene we have an instance of LRRH connected with an identity relationship to the LRRH in the other (real) scene. Then, we have an instance of Grandma which is connected with an identity relationship to the wolf in the real scene. 

Another useful macro statement (\$.//) simplifies repetitive quote statements which have the same quote part but different inquit. Overall, macro statements simplify the repetitive writing of Xapi statements, but do not extend the expressivity of the language.

%
%
\section{Focus and shadows}
\label{sec:FocusAndShadows}

The Xapagy equivalent of a working memory is the {\em focus}, a weighted set of recent instances and VIs. In absence of any events, the weights are gradually decreasing. Instances are reinforced when they participate in new VIs. Action VIs are ``pushed out'' from the focus by their successors, while relation VIs stay in the focus as long as their associated instances are in the focus. While in the focus, instances and VIs can acquire new attributes and relations, and they gradually increase their salience in the autobiographic memory. After an instance or VI leaves the focus, {\em it can never return}. 

The instances and VIs from the autobiographical memory affect the current state of the agent by {\em shadowing} the focus. Each instance and VI in the focus is the {\em head} of a an associated instance or VI set called the {\em body} of the shadow. 

The challenge, of course, is how to populate and maintain the shadows such that they reflect the previous experience of the agent with respect to the ongoing story. The system maintains its internal structures using the interaction between a number of {\em activities}. {\em Spike activities} (SAs) are instantaneous operations, executed one at a time. {\em Diffusion activities} (DAs) represent gradual changes; the output depends on the amount of time the diffusion was running. Multiple DAs run in parallel, reciprocally influencing each other. 

In the following we briefly enumerate the SAs and DAs which maintain the shadows. The + or - prefix indicates whether the activities reinforce or weaken the shadow components. The shadow maintenance activities are {\em self-regulating}, encompassing elements of negative feedback as well as resource limitation. 

\begin{compactitem}

\item[(S+)] The addition of an {\em unexpected} instance or VI creates a corresponding empty shadow.

\item[(S+)] The addition of an {\em expected} instance or VI creates a new shadow from the headless shadow which predicted it.

\item[(D-)] In the absence of other factors, all the shadows decay in time.

\item[(D+)] Instance matching the head: instances from memory which have attributes matching the shadow head will be reinforced in the shadow body. 

\item[(D+)] Instance matching the body: instances from memory which match attributes of some instances from a given shadow, are added to the shadow. This allows shadowing by instances which are indirectly related - i.e. do not have overlapping attributes with the shadow head. 
 
\item[(D+)] VIs whose verb matches the attributes of the shadow head VI are reinforced in the shadow. This process is amplified if the corresponding parts of the VI (subject, object) are already in a shadowing relation. As a side-effect, the shadowing relation of the parts of the VI will be reinforced. 

\item[(D+)] Instances which are in identity relation with the shadow head or any item in the shadow are reinforced in the shadow. 

\item[(D+)] Shadow VIs which have successors / predecessors which are in the shadows of the successor / predecessor of the shadow head are reinforced. Similar DAs apply for the coincidence, context and summarization relations.

\item[(D+/-)] Same-scene instance sharpening: for all the instances in a shadow which share a scene, reinforce the strongest ones and decay the others.

\item[(D+/-)] For all the VIs in a shadow which share the same dominant scene, reinforce the strongest ones and decay the others.

\end{compactitem}

\subsection{Example}

\begin{figure*}
\begin{center}
\includegraphics[scale=0.6]{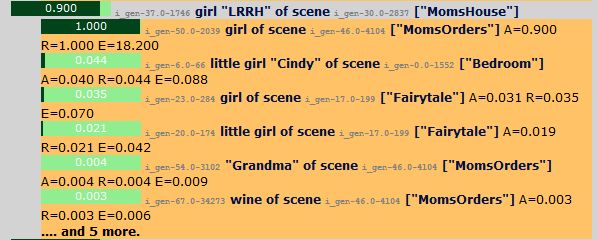} 
\caption{\label{fig:ScreenShadowingLRRH} Screenshot from the web-based GUI of a Xapagy agent. The figure shows
the shadowing of the real instance of LRRH at the moment when she is about to leave to the forest. }
\end{center}
\end{figure*}

Figure~\ref{fig:ScreenShadowingLRRH} shows a cropped screenshot of the shadowing panel of the web-based GUI of the Xapagy agent. The agent was reading the story of LRRH and it was at the point where the girl is about to leave for the forest. The current instance of the girl (from the Momshouse scene) is shadowed by a number of instances. The strongest shadow is the instance of LRRH from the future-fictional scene representing the Mom's orders. However, we also have shadows of other instances of LRRH, other female characters from the story, and also of Cindy, the little girl to which the story is narrated. 

%
%
\section{Headless shadows}
\label{sec:HeadlessShadows}

Headless shadows (HLSs) are collections of related and aligned in-memory VIs which are not paired with any current in-focus VI. The creation and maintenance of HLSs involves three distinct entities: {\em shadow VI relatives (SVRs)}, {\em shadow VI relative interpretations (SVRIs)} and the HLSs themselves. Although for the sake of clarity we describe the creation of these entities in sequential order, these components are maintained by several DAs operating in parallel.

%
%
\subsection{Shadow VI relatives\label{sec:SVR}}

\begin{figure}
\begin{center}
\includegraphics[scale=0.4]{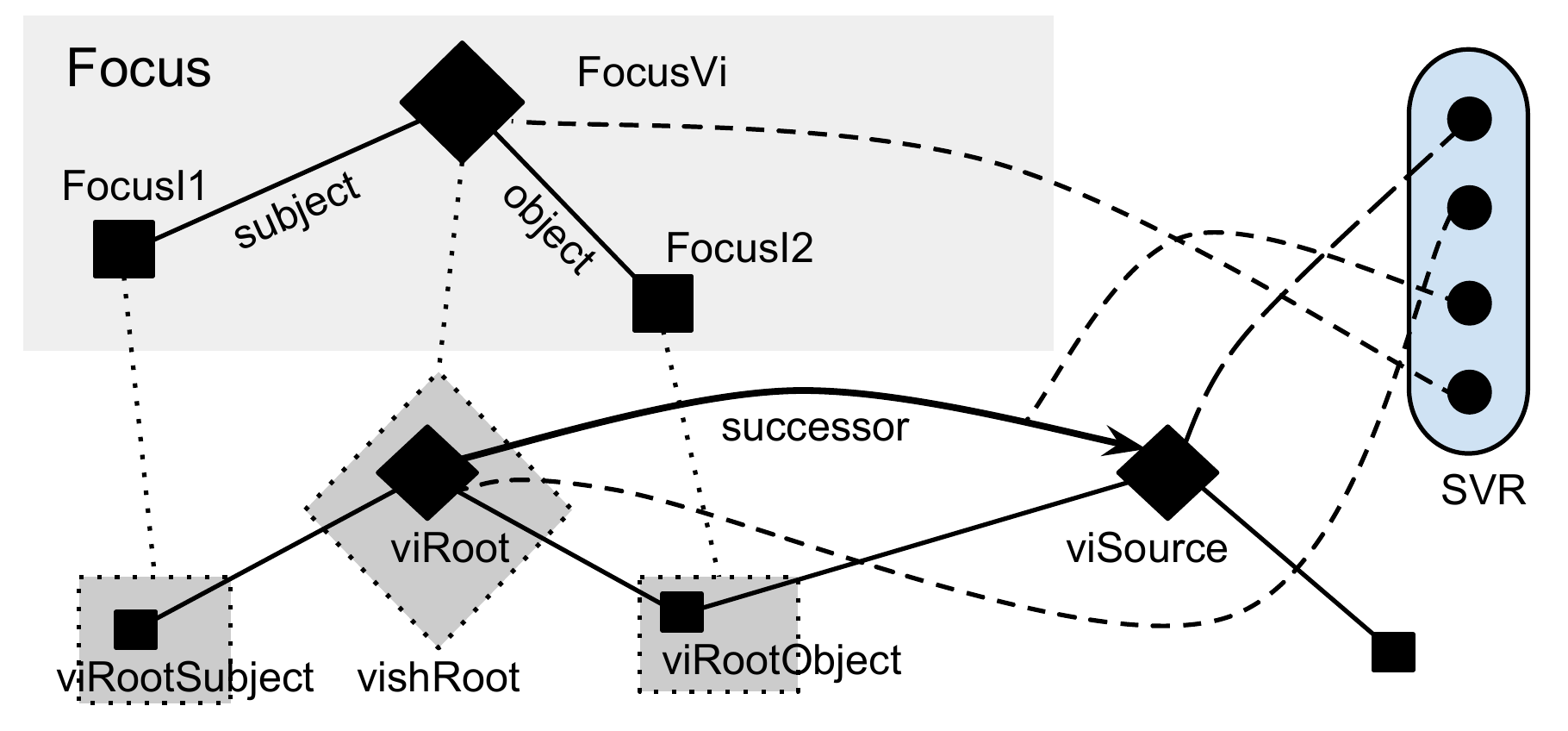} 
\caption{\label{fig:HLS-SVR}The composition of an SVR. VIs are represented as diamonds, instances as squares. Shadows are the same type as the instance they are shadowing.}
\end{center}
\end{figure} 

A shadow VI relative (SVR) is a structure built around a VI which is ``related'' to a VI which is currently in the shadow. Let us consider Figure~\ref{fig:HLS-SVR}. In the focus we have the VI |FocusVI| of type S-V-O, which has its subjects and objects the instances |FocusI1| and |FocusI2| respectively. These elements have their own respective shadows. Let us consider the VI |viRoot| from the shadow of |FocusVI|. This is also of type |S-V-O|, and has its subject and object |viRootSubject| and |viRootObject|, which are part of the shadows of |FocusI1| and |FocusI2| respectively\footnote{These instances can be part of multiple shadows.}. Let us now consider the VI |viSource| which is related to |viRoot| by being connected through a succession relation. The SVR is the quadruplet formed by the |FocusVI|, |viRoot|, |viSource| and the relation type connecting |viRoot| to |viSource|. The latter one is called the {\em type} of the SVR, and it can take nine possible values, all but one arranged in opposing pairs: |IN_SHADOW|, |PREDECESSOR| $\longleftrightarrow$ |SUCCESSOR|, |SUMMARY| $\longleftrightarrow$ |ELABORATION|, |ANSWER| $\longleftrightarrow$ |QUESTION|, |CONTEXT| $\longleftrightarrow$ |CONTEXT_IMPLICATION|.

Intuitively, the |viSource| in the SVR represents a VI which was present when situations similar to the one in the current focus had been encountered. An SVR by itself does not present a prediction with regards to the current focus, because the |viSource| is expressed in terms of in-memory instances, not in-focus instances. In order to find out what kind of prediction does an SVR imply, we must {\em interpret} it.

%
%
\subsection{Shadow VI relative interpretations\label{sec:SVRI}}

\begin{figure}
\begin{center}
\includegraphics[scale=0.4]{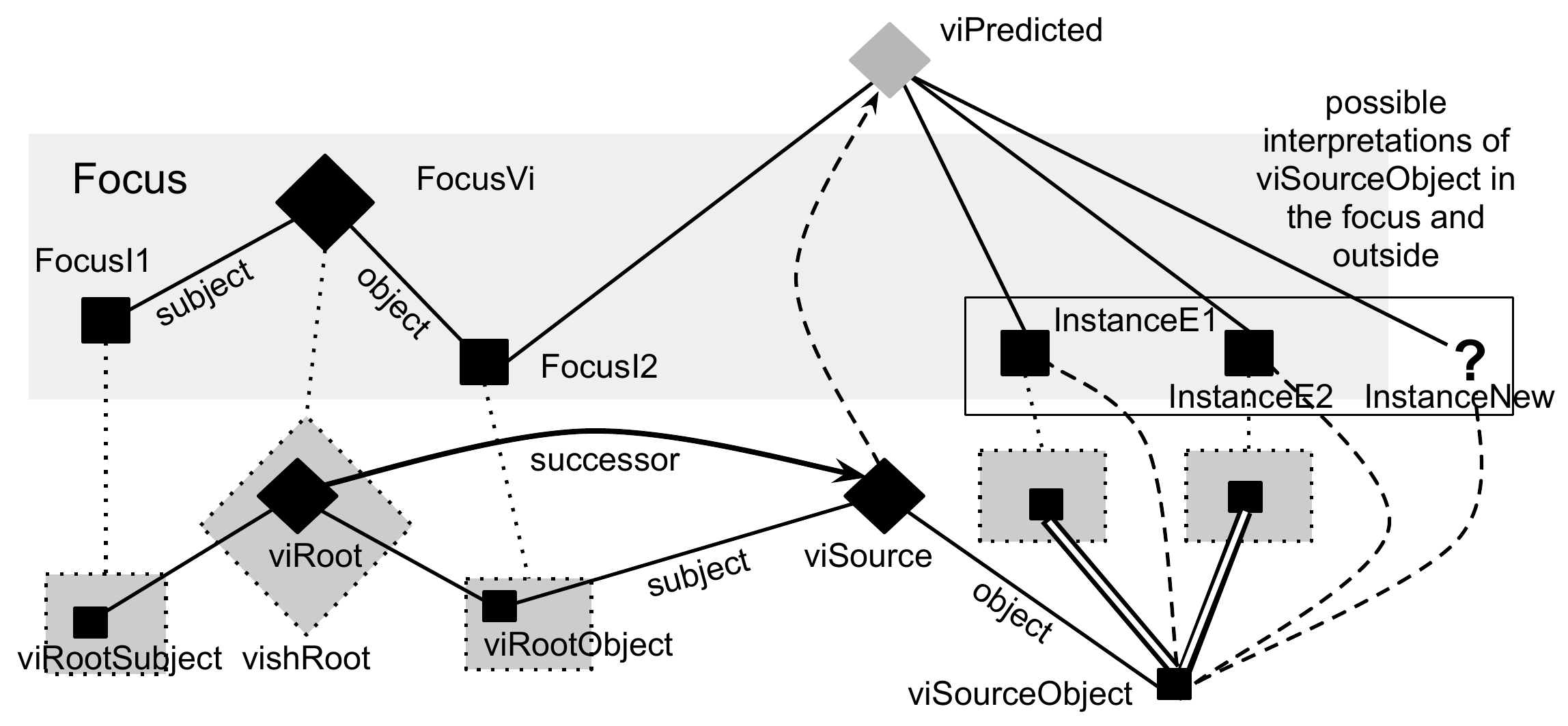} 
\caption{\label{fig:HLS-SVRI} 
Shadow VI relative interpretations.
}
\end{center}
\end{figure} 

Let us take a look at Figure~\ref{fig:HLS-SVRI} which elaborates on Figure~\ref{fig:HLS-SVR}. If |viSource| will be interpreted as a continuation, it will predict the VI |viPredicted| which will also have the format S-V-O. We can make the assumption that the verb in |viPredicted| will be the same as in |viSource|. Finding the subject and the object of |viPredicted| is more complicated: the parts of |viSource| are past instances from the memory which can not be brought back, while the parts of |viPredicted| must be instances in the focus. 

With regards to the subject, we notice that the subject of |viSource| is the same as the object of |viRoot|, which in turn, had been obtained as the shadow of |FocusI2|. We can infer from here than the subject of |viPredicted| will be |FocusI2|. 

The object, however, is more complicated, as it can not be unequivocally inferred from the |viRoot|. Our technique to find an interpretation of |viSourceObject| will be based on ``reverse shadowing'': we look up the shadows in which |viSourceObject| is present, and we consider their heads as candidates with the relative strength of |viSourceObject| in their respective shadows. 

In our case, |viSourceObject| is present in the shadows of |InstanceE1| and |InstanceE2|, these two representing possible interpretations of |viSourceObject|. In addition, there is also a possibility of interpreting |viSourceObject| as an instance which does not yet exist in the focus, |InstanceNew|. 

Putting these considerations together, and denoting the verb of |viSource| with |act|, we have three possible predictions associate with |viPredicted|, which can be described in Xapi as follows:

\begin{quote}
\begin{Verbatim}
The FocusI2 / act / the InstanceE1.
The FocusI2 / act / the InstanceE2.
The FocusI2 / act / an InstanceNew.
\end{Verbatim}
\end{quote}

\noindent where |InstanceNew| is a newly created instance which will be initialized with some of the attributes of |viSourceObject|. These verb instance templates, together with the SVR which are the source of them constitute the SVRIs, weighted by the likelihood of the individual interpretations. 

%
%
\subsection{Headless shadows}

Headless shadows (HLSs) aggregate the support of different types of SVRIs. An HLS is composed of a template for a possibly instantiatable VI and a collection of compatible SVRIs. An SVRI is compatible with a template if the corresponding instance parts are the same and the corresponding concept and verb overlays are ``close''. 

The example used to introduce SVRs and SVRIs was based on a |SUCCESSION| relation, thus we kept referring to them as ``predictions''. The nine different types of SVRIs provide support for or against specific types of HLSs.  Let us consider a case where we see the HLS as a prediction of events to happen next. An SVRI of type |SUCCESSOR| provides evidence that similar events succeeded events in the shadows -- this is a supporting evidence. An SVRI of type |PREDECESSOR| provides evidence that similar events preceded events in the shadow -- which means that they are not successors -- this can be interpreted as a negative evidence. An SVRI of type |IN_SHADOW| means that the given prediction can be mapped back to events which already happened, thus they are not a proper prediction -- again, a negative evidence. An SVRI of type |CONTEXT_IMPLICATION| shows that similar things have happened in similar contexts -- a positive evidence. An SVRI of type |ELABORATION| means that similar things happened when elaborating stories which can be summarized with the same VIs. 

The support of the HLS integrates these evidences into a single number. When an HLS is used to instantiate a new VI, the VI will be created based on the template, while the new shadow will be formed by the |viSource|-s associated with the SVRIs with a positive contribution. After the initial creation of the shadow, this will evolve under the control of the shadow maintenance DAs.

When the HLS is used for a different purpose, the evidences are combined in a different way. For instance, for the inference of a missing action, a |PREDECESSOR| SVRI is a {\em positive} evidence: it can show that we are witnessing the effect of an action which we have not seen.

\subsection{Example}

\begin{figure*}
\begin{center}
\includegraphics[scale=0.6]{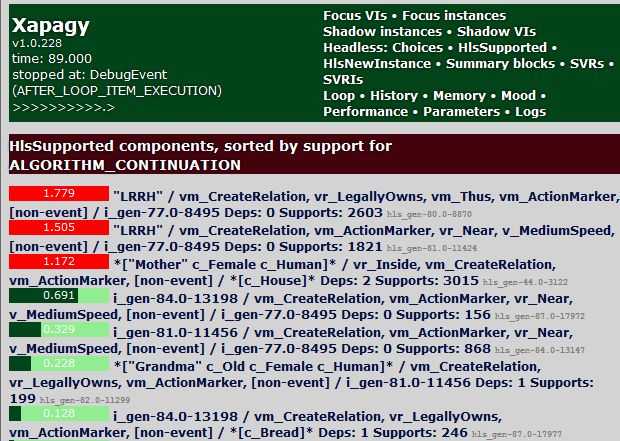} 
\caption{\label{fig:ScreenHLS} Screenshot from the web-based GUI of a Xapagy agent. The figure shows
the headless shadows at the moment when LRRH is about to leave to the forest. }
\end{center}
\end{figure*}

Figure~\ref{fig:ScreenHLS} shows a screenshot of the HLS panel of web-based GUI of the Xapagy agent. The agent was reading the story of LRRH and it was at the point where the girl is about to leave for the forest. The figure shows the HLSs sorted by the support as a continuation. The first (strongest) HLS correctly predicts that LRRH will pick up the basket. Some of the other continuations match the ones what a human observer would predict: that LRRH will go to Granma's house, and Grandma would receive the bread. Other ones are somewhat unexpected: the third predicts that Mother will also be in Grandma's house. 


\section{Narrative reasoning with shadows and HLSs}
\label{sec:NarrativeReasoning}

In the following we shall describe the ways in which the shadow/HLS architecture is used for narrative reasoning, which we will define as the behavior of the agent when it is reading or witnessing certain stories, or when it is recalling or confabulating stories. 

As the Xapagy agent's internal behavior is governed by dynamic processes (in the form of DAs), the behavior is time dependent. For instance, the speed at which new events are delivered to the agent affects the reasoning. Events are instantaneous SAs, while the DAs act in between the arrival of SAs. An agent which has events delivered slowly might be more likely to generalize the story in the form of summarizations, or to infer missing events. On the other hand, if there are significant delays between actions, the agent might have difficulty chaining them in a consistent story. 

The maintenance of the shadows and HLSs in Xapagy is automatic, affected only by the temporal dimension. However,  the narrative reasoning behavior of the agent is also affected by a series of internal variables collected under the term {\em mood}, which balance the willingness of the agent to act on various HLSs by limiting the energy the agent can spend to generate events internally and its preference to various types of events. 

\subsection{Tracking expectedness and surprise}

Incoming events (VIs) are {\em expected} if there is a continuation-type HLS which predicted the VI. The support of the HLS specifies the degree of the expectedness. The predicting HLS will be converted into the shadow of the new VI. 

The addition of a VI not only creates its own shadow, but also affects the shadows of the other VIs and existing instances. We define the measure of this change as the {\em surprise}. For instance, the wolf swallowing LRRH is not unexpected if the agent had seen stories involving large carnivorous animals before - but it is surprising, as it reshuffles the current scene and eliminates a number of possible continuations. 

\subsection{Inferring missing actions}

When tracking the current story, a Xapagy agent might infer missing action HLSs - which differ from continuation HLSs by being supported primarily by predecessor rather than successor links. The Xapagy agent, depending on the mood, might choose to internally instantiate these actions and insert them into the focus. 

Instantiation of missing actions can affect the agent's future behavior. For instance, it commits the agent to a certain interpretation of a story by further reinforcing the shadows from the storyline from which the missing action was itself supported. The inference of the missing action also affects future recalls of the story, as the inferred action becomes an inseparable part of the story.

\subsection{Inferring missing relations}

The inference of the missing relations is based on the missing relation HLSs which are supported by context relation links from the shadows, and inhibited by existing similar or conflicting relations from the focus. For instance, the story might not explicitly specify that LRRH loves her grandma, but the agent, based on a number of stories in its autobiography might infer that she does. 

Just as in the case of missing action, the instantiation of a missing relation affects the future shadowing and expectation tracking of the story. 

We need to emphasize the significant step between inferring an HLS for a missing action or relation, and actually instantiating them. An agent can maintain multiple conflicting HLSs, as long as they are consistent with the focus. However, once an HLS is instantiated, and thus moved into the focus, the DAs will gradually inhibit all the shadows and HLSs which are inconsistent with it. 

\subsection{Summarizations}

Summarizations are VIs which summarize a group of other VIs. They do not appear in the observed flow of the events, but they can appear in the narrated story, and they can be introduced by the agent itself. 

The Xapagy agent performs two types of summarizations. {\em Built-in} summarizations recognize repetitive actions and alternations of actions between two actors. {\em Learned} summarizations are aquired from external narration and can be used to define new verbs. For instance, after the narrator describes a series of hits, cuts and kicks between the hunter and the wolf, he might say:

\begin{quote}
\begin{Verbatim}
The hunter + wolf / in-summary are-fighting.
\end{Verbatim}
\end{quote}
 
After learning the meaning of the |are-fighting| verb through one or more stories like this, the agent will generate summarization HLSs whenever it sees that a similar sequence of events, and, depending of the mood, it might explicitly instantiate the summary. Alternatively, when the agent encounters a summarization verb in a narrative, it can generate HLSs appropriate to its elaboration (and, depending on the mood, instantiate them).
 
\subsection{Recall}

A Xapagy agent can {\em recall} a story by step-by-step instantiating the continuation HLSs. A recall is initialized by bringing the agent in a situation where the continuations point to the appropriate story. Externally, this can be achieved by entering either the recognizable beginning of the story, or a summarization of the story. The agent also must be put in an appropriate recall mood to instantiate the continuations. In a story-following mood, the agent still maintains the continuations, but only uses them to track surprise. 

The instantiated continuation HLSs are not necessarily an exact copy of the recalled story. For instance, the inferred mising actions, relations and summarizations are also part of the recall and they can not be separated from the story. Furthermore, it is very rare that there is a single possible continuation HLS - stories which are similar to the recalled story will also participate in the shadows, affecting the recall. 

Finally, the recalled story is also a story, in the sense that it is a series of instantiated VIs in the autobiographical memory. If an agent had recalled a story many times, those recalls will participate in the shadows. This will strengthen the recall, but it can also distort the recall of the original storyline, as the agent might recall not the original events but the way it had previously narrated them. 


\subsection{Confabulation}

Confabulation and recall are implemented through the same mechanism in Xapagy. The difference is only in the degree of adherence to a certain dominant story line. Confabulation happens when the agent's mood is set in such a way as to relax the rules of shadowing, to allow shadows to contain instances and VIs which are more remotely related. Furthermore, the agent might choose to not instantiate the strongest continuation HLS but to use other criteria for the continuation. This way, an agent generates a new story - which might range from an existing story, slightly modified in the retelling (for instance with a different ending) to a story where every VI is taken from a different story line, effectively creating a new story.


\section{Related work}
\label{sec:RelatedWork}

In the following we will try to position the Xapagy architecture by investigating its relationship to some influential trends in cognitive system design, and its relationship to concrete systems with respect to these trends. 

\medskip

\noindent{\bf The strong-story hypothesis} states that reasoning about stories is a critical part of intelligences~\cite{Winston-2011-StrongStory}. As Xapagy aspires to mimic the cognitive activities humans use in thinking about stories, it naturally subscribed to this view. 

\noindent{\bf The role of worlds:} Many cognitive systems deploy multiple, individually consistent, usually closed models which can represent an interpretation of the present state of the world, a moment in the past, a possible future or an alternate version of reality. These models are often called {\em worlds} or {\em contexts}, although many alternative names exist. For instance, Soar~\cite{lehman1998gentle} dynamically creates structures called {\em substates} whenever it encounters an {\em impasse} in reasoning, and needs new knowledge added to the reasoning pool. In Cyc~\cite{Lenat-1990-Cyc} {\em subtheories} are used to represent alternate versions of reality, for instance, the description of the state of the world in a certain moment in the past (for instance, we can have a microtheory in which Nelson Mandela is still a prisoner). The Polyscheme architecture~\cite{Cassimatis-2004-Integrating} integrates different representations and reasoning algorithms by allowing them to operate over simulated worlds. In \cite{Cassimatis-2009-ReasoningAsSimulation}, the authors show that worlds-based reasoning by simulation can emulate the Davis-Putnam-Logemann-Loveland algorithm and the Gibbs sampling method of probabilistic inference. The {\em scenes} in the Xapagy architecture are an instance of the world model. 
 
\noindent{\bf The role of the autobiographical memory:} Many cognitive systems implement an episodic/ autobiographical memory -- see \cite{Nuxoll-2007-EpisodicMemory} for Soar, and \cite{Stracuzzi-2009-IcarusReasoningOverTime} for ICARUS. However, the importance of the autobiographical memory

\footnote{When referring to the Xapagy system, we prefer to use the term ``autobiographical memory'' rather than ``episodic memory''. The latter is strongly associated with the work of Tulving. However, in Tulving's view episodic memory is a ``recently evolved, late-developing [\ldots] past-oriented memory system'' whose ``operations require, but go beyond, the semantic memory system'' \cite{Tulving-2002-FromMindToBrain}. In contrast, the autobiographical memory in Xapagy is not the culmination, but the foundation of all other memory-like behaviors.} for Xapagy is more critical. The system has no procedural or skill memory, no rule or production memory, and no concept hierarchy. The content of the working memory (the {\em focus}) can be moved into the long term (autobiographical) memory, but not the other way around. The agent cannot reload a previous experience, nor parts of it. The autobiographical memory influences the behavior of the agent only through the shadowing mechanism. The content of the autobiographical memory is {\em never} extracted into general purpose rules: there is no learning, only a recording of the experiences.

\noindent{\bf Common serial mechanism.} Xapagy makes the assumption that acting, witnessing, story following, recall and confabulation are implemented by a common serial mechanism. A number of other cognitive architectures make the same assumption, for instance, ACT-R~\cite{Anderson-1998-AtomicComponentsOfThought,Anderson-2004-IntegratedTheoryOfMind}. Combined with the other design decisions of Xapagy, however, this triggers several unexpected implications. The first is the {\em undifferentiated representation of direct and indirect experiences}. The stories exiting from the story bottleneck are recorded together in the autobiographical memory, with no fundamental distinguishing feature. The second implication is the {\em unremarkable self}. The Xapagy agent maintains an internal representation of its cognition (the real-time self), in the form of an instance labeled |"Me"|. However, this instance is not fundamentally different from the instances representing other entities. Together with the inability to recall instances from the memory, this yields another implication, the {\em fragmentation of the self}. As the entity of the self can not be retrieved from memory, only recreated, an agent remembering its own stories will have simultaneously several representations of itself, only one of them marked as its real time self. Thus, {\em every recall of a story creates a new story}. 

\noindent{\bf Handling questions.} As a system reasoning about stories, Xapagy needs to deal with the problem of questions - both in terms of questions appearing in the dialogs of the stories, as well as possible questions about the story which might be answered by the agent. Recent work in statistical NLP had successfully demonstrated answering questions by parsing large databases~\cite{Ferrucci-2010-BuildingWatson}. 

The Xapagy system's approach is more closely related to the structural question answering models, where the answers are provided from the autobiographical knowledge of the agent, rather than from the parsing of large databases. In particular, the Xapagy approach is more closely related to that of the AQUA project of Ashwin Ram~\cite{Ram-1991-TheoryOfQuestions,Ram-1994-AQUA}. The main difference is that in AQUA the ongoing assumption is that there is a schema based knowledge model which is enhanced through question driven learning. There is no such model in Xapagy. Questions and answers become part of the autobiographical memory, but there are no higher level knowledge structures.
AQUA seeks truthful answers to questions externally posed, and raises internal questions for seeking knowledge. In Xapagy questions are simply a specific type of sentence and they can represent a number of scenarios: questions for which an answer is sought, rhetorical questions, questions whose goal is to find out if the interlocutor knows something, questions asked to influence the discourse and so on. There is no strong tie between the question and learning, and there is no assumption with regards to the fact that a question must be answered, and if yes, truthfully. The questions of Little Red Riding Hood and the wolf's answers are an illustrative case. 

\noindent{\bf Relationship to logic-based approaches.} Reasoning about stories had been successfully demonstrated using logic-based approaches. Situation calculus, event calculus and action languages had all been adapted to story understanding and question answering. For instance, Eric Mueller had applied event calculus to the understanding of script based stories of terrorist incidents~\cite{Mueller-2004-Understanding}. 

Another approach for representing stories using logic is the {\em episodic logic} proposed by Lenhart K. Schubert. Episodic logic allows the translation of the English language stories into a rich logical model. In \cite{Schubert-2000-EpisodicLogicLRRH} a portion of the story of LRRH is translated into EL. This covers not only basic narrative facts, such as ``who did what to whom'', but very sophisticated sentences, for instance: 

\begin{quote}
{\em The wolf would have very much liked to eat [LRRH], but he dared not do so on account of some woodcutters nearby. }
\end{quote}

The challenge of the use of episodic logic is that it requires a significant amount of knowledge engineering of the background rules. The authors state as a general principle that the knowledge must be specified at the most general level possible. An example of this would be rules such as ``if a predator encounters a non-predatory creature not larger than himself and it is enraged or hungry, it will try to attack, subdue and eat that creature''.

The same problem, of course, appears in Xapagy as well, but in a different way: instead of background rules, the system requires relevant autobiographical knowledge. The history of the Xapagy agent must contain experiences with predators, which will shadow the current story. 

\noindent{\bf Learning from reading.} Xapagy acquires most of its knowledge by reading narratives. One of the recent approaches which acquire knowledge by reading is the work led by Kenneth Forbus in the DARPA sponsored ``learning by reading'' project at Northwestern University~\cite{Forbus-2007-LearnByReading}. For instance, the system is able to learn about the history of the Middle East by reading historical narratives and newspaper reports and it can answer specific questions about the problem domain. The system reads texts in simplified English and relies on the Direct Memory Access Parsing (DMAP) model~\cite{Martin-1986-Uniform} which tightly integrates the processing of the natural language with the knowledge representation and reasoning. Background knowledge is provided by the ResearchCyc database. 

\section{Conclusions and future work}
\label{sec:Conclusions}

This paper described the narrative reasoning technique of the Xapagy architecture. The system differentiates itself from other cognitive architectures through its exclusive reliance on the raw autobiographical memory and its fragmentation of real world entities into loosely interconnected instances. Accordingly, the reasoning techniques of the system are adapted to these features. The shadowing technique aligns relevant experiences from the autobiography with the currently witnessed events stored in its focus. The shadows, which are attached to recent observations, are extrapolated into headless shadows (HLSs). HLSs are used to track the expectations and level of surprise of the agent by predicting the continuation of an unfolding story, to infer missing actions or missing relations, and to summarize recent experiences. In recall mood, HLSs can be used to generate new stories, either by more or less exactly recalling a previous story-line, or by confabulating new stories based on fragments of previous stories.

\bibliography{../../Bibliography/Xapagy,../../Bibliography/ClassicStoryUnderstanding,../../Bibliography/CognitiveArchitectures,../../Bibliography/ComputationalNarrative,../../Bibliography/PhilosophyPsychology}
\bibliographystyle{abbrv} 

\end{document}